\newif\ifarxiv
\arxivtrue 

\ifarxiv
\documentclass{article}
\usepackage[margin=1in]{geometry}
\else
\documentclass[conference]{IEEEtran}
\fi

\usepackage{graphicx}
\usepackage{float}
\usepackage{amsmath}
\usepackage{amssymb}
\usepackage{algorithmic}
\usepackage{array}
\usepackage{color,soul,marginnote}

\DeclareMathOperator*{\mini}{minimize}

\usepackage{url}

\begin{document}
\title{Deep Reinforcement Learning for Dynamic Urban Transportation Problems}

\ifarxiv
\author{Laura Schultz and Vadim Sokolov\\
	   Systems Engineering and Operations Research\\
		George Mason University. Fairfax, VA\\
		Email: \{lschult2, vsokolov\}@gmu.edu}
\date{First Draft: May 2018\\
	This Draft: June 2018.}
\else
\author{\IEEEauthorblockN{Laura Schultz and Vadim Sokolov}
\IEEEauthorblockA{Systems Engineering and Operations Research\\
George Mason University. Fairfax, VA\\
Email: \{lschult2, vsokolov\}@gmu.edu}
%
%
%
}
\fi

%


\maketitle
\begin{abstract}
We explore the use of deep learning and deep reinforcement learning for optimization problems in transportation. Many transportation system analysis tasks are formulated as an optimization problem - such as optimal control problems in intelligent transportation systems and long term urban planning. Often transportation models used to represent dynamics of a transportation system involve large data sets with complex input-output interactions and are difficult to use in the context of optimization. Use of deep learning metamodels can produce a lower dimensional representation of those relations and allow to implement optimization and reinforcement learning algorithms in an efficient manner. In particular, we develop deep learning models for calibrating transportation simulators and for reinforcement learning to solve the problem of optimal scheduling of travelers on the network.
\end{abstract}

%
\ifarxiv
\else
\IEEEpeerreviewmaketitle
\fi

\section{Introduction}
\label{Introduction}
Many modern transportation system analysis problems lead to high-dimensional and highly nonlinear optimization problems. Examples include fleet management~\cite{nair2011fleet}, intelligent system operations~\cite{lam2016learning} and long-term urban planning~\cite{spiess1989optimal}. Transportation system models used in those optimization problems typically assume analytical formulations using mathematical programming~\cite{larson2016coordinated,sokolov2017maximization} or conservation laws~\cite{claudel2010lax,polson2015bayesian} and rely on high level abstractions, such as origin-destination matrices for demand. An alternative approach is to model transportation system using a  complex simulator~\cite{nagel_agent-based_2012,sokolov2012flexible,auld2016polaris}, which 
model individual travelers and provide flexible approach to represent traffic and demand patterns in large scale multi-modal transportation systems. However, solving optimization problems and dynamic-control problems that rely on simulator models of the system is prohibitive due to computational costs. Recently matamodel based approach was proposed  to solve simulation-based transportation optimization problems~\cite{chong2017simulation,schultz2018bayesian}. 

In this paper, we propose an alternative approach to solve optimization problems for large scale transportation systems. Our approach relies on low complexity metamodels deduced from complex simulators and reinforcement learning to solve optimization problems. We use deep learning approximators for the low complexity metamodels.  A deep learning is a Latent Variable Model (LVM) that is capable of extracting the underlying low-dimensional pattern in a high-dimensional input-output relations. Deep learners have proven highly effective in combination with Reinforcement and Active Learning~\cite{mnih2015human} to recognize such patterns for exploitation. Our approach builds on the work on simulation-based optimization~\cite{schultz2018bayesian,chong2017simulation}, deep learning~\cite{polson2017deep,dixon2017deep} as well as reinforcement learning~\cite{wu2017framework,wu2017flow} techniques recently proposed for transportation applications. The two main contribution of this paper are
\begin{enumerate}
	\item  Development of innovative deep learning architecture for reducing dimensionality of search space and modeling relations between transportation simulator inputs (travel behavior parameters, traffic network characteristics) and outputs (mobility patterns, traffic congestion)
	\item Development of reinforcement learning techniques that rely on deep learning approximators to solve optimization problems for dynamic transportation systems
\end{enumerate}

We demonstrate our methodologies using two applications. First, we solve the problem of calibrating a complex, stochastic transportation simulators which need to be systematically adjusted to match field data. The problem of calibrating a simulator is the key for making it useful for both short term operational decisions and long term urban planning. We improve on the previously proposed
numerous approaches for the calibration of simulation-based traffic flow models have been produced by treating the problem as an optimization issue \cite{cheu_calibration_1998, ma_genetic_2002,lu_enhanced_2015, cipriani_gradient_2011, lee_new_2009,hale_optimization-based_2015,hazelton_statistical_2008,flotterod_general_2010,flotterod_bayesian_2011,djukic_efficient_2012}. Our approach makes no assumption about the form of the simulator and types of inputs and outputs used. Further, we show that deep learning models are more sample efficient when compared to Bayesian techniques or more traditional filtering techniques.  We build on our calibration framework~\cite{schultz2018bayesian} by further exploring the dimensionality reduction utilized for more efficient input parameter space exploration. More specifically, we introduce the formulation and analysis of a combinatorial Neural Network method and compare it with previous work that used Active Subspace methods.

The second application builds upon recent advances in deep learning approaches to reinforcement learning (RL) that have demonstrated impressive results in game playing \cite{mnih2013playing} through the application of neural networks for approximating state-action functions.  Reinforcement Leaning mimics the way humans learn new tasks and behavioral policies via trial and error and has proven successful~\cite{sutton2017} in many applications. While most of the research on RL is done in the field of machine learning and applied to classical AI problems, such as robotics, language translation and supply chain management problems~\cite{giannoccaro2002inventory}, some classical transportation control problems have been previously solved using RL. \cite{abdulhai2003reinforcement,arentze2000albatross,bingham2001reinforcement,bazzan2009opportunities,abdulhai2003reinforcement,arel2010reinforcement,ling2005reinforcement,cunningham2008collaborative, adam_evaluating_2009, chong_revised_2011}.  Furthermore, there were recent attempts that successfully demonstrated applications of deep RL to traffic flow control ~\cite{adam2009evaluating,wu2017framework,wu2017flow,belletti2017expert,genders2016using}. 

The remainder of this paper is organized as follows: Section~\ref{sec:dl} briefly documents the highlights of neural network architectures; Section~\ref{sec:calibration} describes the new deep learning architecture that finds low dimensional patterns in simulator's  inputs-output relations and we apply our deep learner to the problem of model calibration. Section~\ref{sec:rl} describes the additional application of deep reinforcement learning to transportation system optimization. Finally Section~\ref{sec:discussion} offers avenues for further research.

\section{Deep Learning}
\label{sec:dl}
Let $y$ denote a (low dimensional) output and $ x = (x_1 , \ldots , x_p ) \in \Re^p $  a (high dimensional) set of inputs.  We wish to recover the multivariate function (map), denoted by $ y = f( x) $, using training data of input-output pairs $( x_i , f( x_i ) )_{i=1}^N $, that generalizes well for out-of-sample data. Deep learning uses compositions of functions, rather than traditional additive ones. By composing $L$ layers, a deep learning predictor becomes
\begin{align*}
\hat y = & F_{w,b}(x) = (f_{w_0,b_0} \circ \ldots f_{w_L,b_L})(x_1,\ldots,x_p)\\
f^l_{w_l,b_l} = & f_l(w_lx_l +b_l).
\end{align*}
Here $f_l$ is a univariate activation function. The set $(w,b) = (w_1,\ldots,w_L,b_1,\ldots,b_L)$ is the set of weights and offsets which are learned from training data. Here $w_l \in R^{p_l \times p_l'}$ and dimensionality $p_l$ is of the architecture specifications. 

Training the parameters $(W,b)$ and selecting an architecture is achieved by regularized least squares. Stochastic gradient descent (SGD) and its variants are used to find the solution to the optimization problem~\cite{polson2017deep}.
$$
\mini_{W,b} \; \sum_{i=1}^N ||Y_i - F_{W,b} ( x_i ) ||_2^2 + \phi ( W,b ) ,
$$
where $(Y_i,x_i)_{i=1}^N$ is training data of input-output pairs, and $ \phi( f_{\text{DL}} ) $ is a regularisation penalty on the network parameters (weights and offsets).

In this paper we develop a new deep learning architecture for simultaneously learning  low dimensional representation of simulator's input parameter space as well as the relation between simulator's inputs and outputs. Our architecture relies on multi-layer perceptron and auto-encoding layers. Multilayer Perceptron Network (MLP) -- a neural network which takes a set of inputs, $x = (x_1, x_2,...x_i)^T$, and feeds them through one or more sets of intermediate layers to compute one or more output values, $Y = (y_1, y_2,...y_n)^T$. Although a single architecture is commonly implemented in practice, some success has been found through comparative or combinatorial means\cite{zhang_nonlinear_2001}.

\subsection{Auto-Encoder} 
An auto-encoder is a deep learning routine which trains the architecture to approximate $x$ by itself (i.e., $x$ = $y$) via a bottleneck structure. This means we select a model $F_{W,b}(x)$ which aims to concentrate the information required to recreate $x$. Put differently, an auto-encoder creates a more cost effective representation of $x$. For example, under an L2-loss function, we wish to solve 
\[\mini_{W,B} ||F_{W,b}(x)-x||_2^2\]
subject to a regularization penalty on the weights and offsets. In an auto-encoder, for a training data set $\{x_1,\ldots,x_n\}$, we set the target values as $y_i = x_i$. A static auto-encoder with two linear layers, akin to a traditional factor model, can be written as a deep learner as
\begin{align*}
a^{(1)} = &x\\
a^{(2)} = &f_1(w^{(2)}a^{(1)} + b^{(2)})\\
a^{(3)} = & F_{W,b}(x) = f_2(w^{(3)}a^{(2)} + b^{(3)}),
\end{align*}
where $a^{(i)}$ are activation vectors. The goal is to wind the weights and biases so that size of $w^{(2)}$ us much smaller than size of $w^{(3)}$

\section{Deep Learning for Calibration}\label{sec:calibration}
In this section we develop a new deep learning architecture that can be used to learn low-dimensional structure in the simulator input-output relations. Then we use the low dimensional representations to solve an optimization problem. Our optimization problem is the problem of  calibration of a transportation simulator. As the simulators become more detailed, the high dimensionality has now become a pressing concern. In practice, high dimensional data possesses a natural structure within it, that can be expressed in low dimensions. Known as Dimension Reduction, the effective number of parameters in the model reduces and enables analysis from smaller data sets.

Recently, we developed a low-dimensional metamodel for to be used for transportation simulation-based problems~\cite{schultz2018bayesian}. We calculated active subspaces to capture low-dimensional structures in the simulator. We used Gaussian process to  represent the input-output relations.  There are several key concerns while developing an efficient simulation-based algorithms for transportation applications

%
%
%

\begin{enumerate}
	\item Algorithms must be sample efficient and parallelizable. Each simulation run is computationally expensive and can take up to a few days. This computational constraint could potentially limit the scale and scope of calibration investigations and result in large areas of sample space unexplored and sub-optimal decisions. As High-Performance Computing (HPC) resources have become increasingly available in most research environments, new modes of computational processing and experimentation have become possible -- parallel tasking capabilities allow multiple simulated runs to be performed simultaneously and HPC programs aid in coordinating worker units to run codes across multiple processors to maximize the available resources and time management. By leveraging these advances and running a queue of pending input sets concurrently through the simulator, a larger set of unknown inputs can be evaluated in an acceptable time frame.	
	
	The variational landscape for a simulation model will not be uniform throughout the state-space. Although active conservation of restricted resource allocations can be mitigated by HPC, additional care should be taken to determine when exploration or exploitation should be encouraged given the data collected and redundant sampling avoided.
	
	A machine learning technique known as \textit{active learning} is leveraged to provide such a scheme. A utility function (a.k.a acquisition function) is built to balance the exploration of unknown portions of the input sample space with the exploitation of all information gathered by the previous evaluated data, resulting in a prioritized ordering reflecting the motivation and objectives behind the calibration effort.	The expectation of the utility function is taken over the Bayesian posterior distribution and maximized to provide a predetermined number of recommendations.
	
	\item Transportation modelers have many ways to model complex interactions represented within the transportation simulator. Calibration methodologies that account for internal structure of a simulator could be more efficient. On the other hand, they are hardly generalizable to other types of simulation models. By treating the relationship between the inputs and outputs of the simulator in question as an unknown, stochastic function, black-box optimization methodologies can be leveraged. Specifically, our previously developed Gaussian process framework  took the Bayesian approach to construct a probability distribution over all potential linear and nonlinear functions matching the simulator and leveraging evidential data to determine the most likely match. Once this distribution is sufficiently mapped, a valued estimation for the sought parameters can be made with minimal uncertainty.
\end{enumerate}

\subsection{Deep Learning Architecture}\label{Network_Config}

Within the calibration framework, two objectives must be realized by the neural network:
\begin{enumerate}
	\item A reduced dimension subspace which captures the relationship between the simulator inputs and outputs must be bounded in order for adequate exploration 
	\item Given the reduced dimension sample determined by the framework, a method to convert the reduction to the original dimension subspace must exist to allow for simulator evaluations
\end{enumerate} 

\begin{figure}[]
	\centering
	\includegraphics[width=0.6\linewidth]{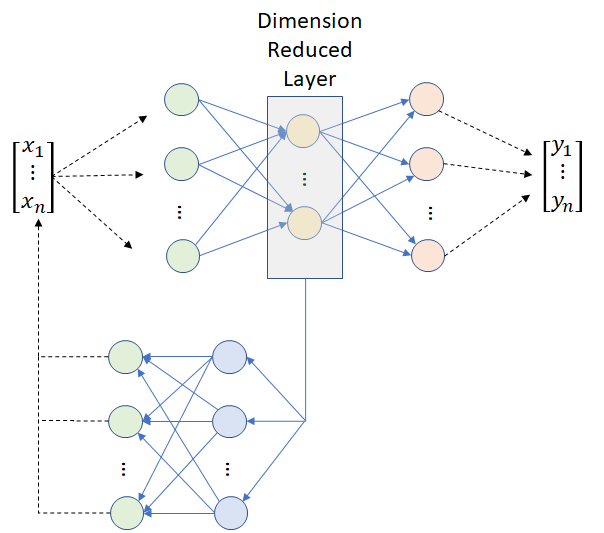}
	\caption{Graphical Representation of the Combinatorial Neural Network for Calibration}
	\label{fig:MLPAuto}
\end{figure}

To address these objectives, we use  MLP architecture to capture the relations between inputs and outputs. We use an coder architecture to capture the low dimensional structure in the input parameters. We will run optimization algorithms inside the low dimensional representation of the input parameter space to address the curse of dimensionality. The Autoencoder and MLP share the same initial layers up to the reduced dimension layer, as shown in Figure \ref{fig:MLPAuto}.

The activation function used is the $\tanh$, which has a range of $(-1,1)$ and is a close approximation to the sign function. 

We ran the simulator several times to generate initial sample set was used by the calibration framework to explore the relationship between the inputs and outputs. Additionally, to quantify the discrepancies during training, the following loss functions were used:
\begin{enumerate}
	\item The MLP portion of the architecture used the mean squared error function $L$
	\begin{equation}
	\displaystyle L\left[y,\phi(\theta)\right] = \frac{1}{N}\sum_{i=1}^{N}(y - \phi(\theta_i))^2
	\end{equation}
	where $\phi(\theta)$ represents the predicted values produced by the neural network for the simulator's output $y$ given the input set $\theta$
	\item The Autoencoder portion of the architecture used the mean squared error function $L$ and a quadratic penalty cost $P$ for producing predicted values outside of the original subspace bounds
	\begin{equation}
	\displaystyle D\left[\phi(\theta)\right] = \max[0,\phi(\theta_i)-x_u]^2 + \max[0,x_l-\phi(\theta_i)]^2
	\end{equation}
	where $\phi(\theta)$ represents the predicted values produced by the neural network for the simulator's input $x$ given the input set $\theta$, $x_u$ represents the input set's upper bound, and $x_l$ represents the input set's lower 
\end{enumerate}

 \subsection{Empirical Results}
 \label{Empirical1}

We use Sioux-Falls~\cite{leblanc_efficient_1975}, transportation model for our empirical results. This model consists of 24 traffic analysis zones and 24 intersections with 76 directional roads, or arcs. The network structure and input data provided by~\cite{stabler_transportationnetworks:_2017} have been adjusted from the original dataset to approximate hourly demand flows in the form of Origin-Destination (O-D) pairs, the simulation's input set.
 
The input data is provided to a simulator package which implements the iterative Frank-Wolfe method to determine the traffic equilibrium flows and outputs average travel times across each arc. 
Due to limited computing availability, only the first twenty O-D pairs are treated as unknown input variables between $0$ and $7000$ which need to be calibrated while the other O-D pairs are assumed to be known. Random noise is added to the simulator to emulate the observational and variational errors expected in real-world applications. 
The calibration framework's objective function is to minimize the mean discrepancy between the simulated travel times resulting from the calibrated O-D pairs and the 'true' times resulting from the full set of true O-D pair values.

\begin{figure}[]
	\begin{tabular}{cc}
		\includegraphics[width=0.45\linewidth]{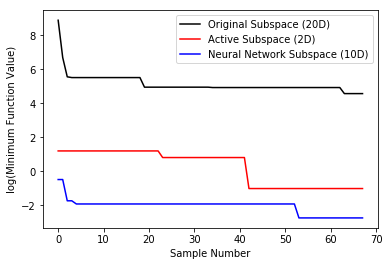} &
		\hspace*{0mm}\includegraphics[width=0.45\linewidth]{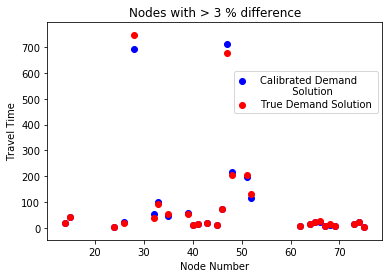}\\
		(a) Objective Function  & (b) Calibrated vs True \\	
	\end{tabular}
	\caption{ Results of demand matrix calibration using Bayesian optimization. (a) Comparison of the Three Methods in terms of Objective Evaluations applied to original parameter space (black line), reduced dimensionality parameter space of Active Subspaces (red line), and reduced dimensionality parameter space of Neural Networks(blue line). (b) Comparison of Calibrated and True Travel Time Outputs with Above Average Differences.}
	\label{fig:results}
\end{figure}

Overall, the performance of the calibration using a deep neural network proved significant, see Figure \ref{fig:results}(a). A calibrated solution set was produced which resulted in outputs, on average, within 3\% of the experiment's true output. With a standard deviation of 5\%, Figure \ref{fig:results}(b) provides a visualization for those links which possessed greater than average variation from the true demand's output. Given the same computational budget, Bayesian optimization that uses low dimensional representation from the deep learner leads to 25\% more accurate match between measured and simulated data when compared to active subspaces.

\section{Deep Reinforcement Learning}
\label{sec:rl}
Consider the desire for a calibrated simulator not to be used for the evaluation of interested scenarios but as a tool for designing a policy $\pi: s \rightarrow a$ which dictates an optimal action $a$ for the current state of the system $s$.

Once calibrated, the simulator is no longer regarded as a black-box but as an interactive system of players, known as agents, and their environment. In such a system, the agent interacts with an environment in discrete time steps. At each timestep, $t$, the agent has a set of actions,$a_{1,\ldots,n}$ which can be executed. Given the action, the environment changes from its original state, $s$, to a new, influenced state, $s'$. If, for every action or through a set of actions, a reward is derived by the agent, the sequential decision problems can be solved through a concept known as Reinforcement Learning (RL).

Such a structure is quite conducive to transportation. For example, if a commuter chooses to leave the house after rush hour has ended, he will eventually be rewarded at the end of his commute with a shorter travel time than if he had left at the beginning of rush hour. Although not immediately realized, the reward is no less desired and will, in the future, encourage the agent to perform the same behavior when possible.

The quantification of such an action-reward trade-off is represented through a function known as 'Q-learning'\cite{watkins_q-learning_1992}. Q-learning, referencing the 'quality' of a certain action within the environment's current state, represents the maximum, discounted reward that can be obtained in the future if action $a$ is performed in state $s$ and all subsequent actions are continued following the optimal policy $\pi$ from that state on:
\[Q_\pi(s,a) = \mathbb{E}[R_t | s_t = s, a_t = a, \pi]\]

where $R_t = \sum_{\tau = t}^{\infty} \gamma^{\tau-t}r_\tau$ is the discounted return and $\tau \in [0,1]$ is the factor used to enumerate the importance of immediate and future rewards.

In other words, it is the greatest reward we can expect given we follow the best set of action sequences after performing action $a$ in state $s$. Subsequently, the optimal policy requires choosing the optimal, or maximum, value for each state:
\[Q^\ast(s,a) = \max_\pi Q_\pi(s,a)\] 

While most of the research on RL is done in the field of machine learning and applied to classical AI problems, such as robotics, language translation and supply chain management problems \cite{giannoccaro2002inventory}, some classical transportation control problems have been previously solved using RL. \cite{abdulhai2003reinforcement,arentze2000albatross,bingham2001reinforcement,bazzan2009opportunities,abdulhai2003reinforcement,arel2010reinforcement,ling2005reinforcement,cunningham2008collaborative, adam_evaluating_2009, chong_revised_2011}. 

Unfortunately, the Q-functions for these transportation simulators continue to possess high-dimensionality concerns noted in our previous calibration work. However, recent advancements have allowed for the successful integration of reinforcement learning's Q-function with deep neural networks.\cite{mnih_human-level_2015} Known as a Deep Q Network (DQN), these neural networks have the potential to provide a diminished feature set for highly structured, highly-dimensional data without hindering the power of the reinforcement learning.

For development and training of such a network, a neural network architecture best-fitting the problem is constructed with the following loss function\cite{wang_dueling_2015}

\[ L_i(\theta_i) = E_{s,a,r,s'} \left[(y_i^{DQN} - Q(s,a; \theta_i))^2\right] \]

where $\theta$ are the parameters, $Q(\cdot)$ is the Q-function for state $s$ and action $a$ and

\[y_i^{DQN} = r + \gamma \max_{a'} Q(s',a'; \theta^-)\]

where $\theta^- $ represents parameters of a fixed and separate target network. 

Furthermore, to increase the data efficiency and reduce the correlation among samples during training, DQNs leverage a buffer system known as \textit{experience replay}. Each transition and answer set, $(s_t,a_t,r_t,s_{t+1})$, is stored offline and, throughout the training process, random small batches from the replay memory are used instead of the most recent transition.

For the purpose of this paper, a MLP network is utilized as the neural architecture.

\subsection{Empirical Results}

\begin{figure}[]
	\centering
	\includegraphics[width=0.3\linewidth]{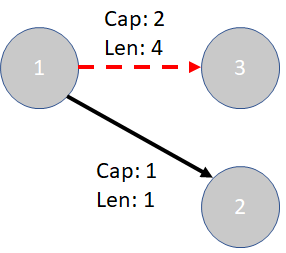}
	\caption{Graphical Representation of the Example Transportation System}
	\label{fig:MLPSys}
\end{figure}

For demonstration and analysis, a small transportation network, depicted in Figure \ref{fig:MLPSys}, consisting of 3 nodes and 2 routes, or arcs, is used.

The small network has varying demand originating from node $1$ to node $2$ for $24$ time periods. Using RL, we find the best policy to handle this demand with the lowest system travel time given that any single period has two allowable actions:
\begin{enumerate}
	\item $0-2$ units of demand from node $1$ to node $2$ can be delayed up to one hour
	\item $0-2$ units of demand from node $1$ to node $2$ can be re-routed to node $3$ as an alternative destination at a further distance
\end{enumerate}
In essence we solve the optimal traffic assignment problem. Our state contains the following information: (i) the amount of original demand from node $1$ to node $2$ that is to be executed at time $t$, $D_{t,1,2}$; (ii) the amount of demand moved to time $t$ for execution from time $t-1$, $M_{t,1,2}$; (iii) the amount of demand left to be met between time $t+1$ and $t=24$ divided by the amount of time left $24-(t+1)$ The action includes the option to move $0$,$1$,or $2$ units of demand from the current period $t$ to the subsequent period $t+1$ or move $0$,$1$,or $2$ units of demand from the arc between note $1$ and node $2$ to the arc between node $1$ and node $3$ , $A_{t} = [A_{t,t+1,1,2},A_{t,1,3}]$. The reward is calculated using the same simulator package from the Section \ref{Emperical1},which implements the iterative Frank-Wolfe method to determine the traffic equilibrium flows and outputs total system travel time for the period. Since Q-learning seeks the maximum reward, we took the negative total system travel time.

\begin{figure}[H]
	\centering
	\includegraphics[width=0.7\linewidth]{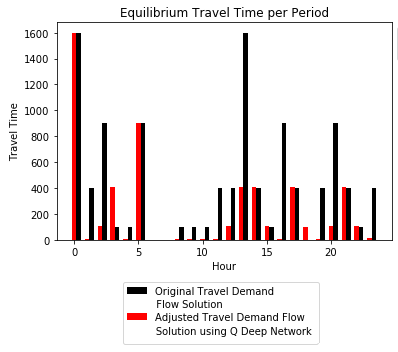}
	\caption{Comparison of System Travel Time per Period}
	\label{fig:QDiff}
\end{figure}

After running the network on 100 of $24$-long episodes, a randomly generated set of demand was produced and run through the resulting neural network. A $51\%$ improvement in the system travel time was achieved. Table \ref{table:add} illustrates the adjustments decided by the $Q$ network and Figure \ref{fig:QDiff} compares the travel times by period between the original and adjusted demands.

\begin{table}[H]
	\caption{Adjustments to Demands per Period using DQN}
	\label{table:add}
	\begin{center}
		\begin{tabular}{p{0.05\linewidth}|p{0.15\linewidth}|p{0.15\linewidth}|p{0.15\linewidth}|}
			
			\textbf{Hour} & \textbf{Original Demand of $Arc_{1,2}$} & \textbf{DQN Adjusted Demand of $Arc_{1,2}$} & \textbf{DQN Adjusted Demand of $Arc_{1,3}$} \\\hline
			
			1&	4&	4&	0 \\
			2&	2&	0&	1 \\
			
			3&	3&	1&	1 \\
			
			4&	1&	2&	1 \\
			
			5&	1&	0&	1 \\
			
			6&	3&	3&	0 \\
			
			7&	0&	0&	0 \\
			
			8&	0&	0&	0 \\
			
			9&	1&	0&	1 \\
			
			10&	1&	0&	1 \\
			
			11&	1&	0&	1 \\
			
			12&	2&	0&	1 \\
			
			13&	2&	1&	1 \\
			
			14&	4&	2&	1 \\
			
			15&	2&	2&	1 \\
			
			16&	1&	1&	1 \\
			
			17&	3&	0&	1 \\
			
			18&	2&	2&	1 \\
			
			19&	0&	1&	0 \\
			
			20&	2&	0&	1 \\
			
			21&	3&	1&	1 \\
			
			22&	2&	2&	1 \\
			
			23&	1&	1&	1 \\
			
			24&	2&	0&	2 \\
			
		\end{tabular} 	
	\end{center}
\end{table}

\section{Discussion}\label{sec:discussion}
Deep learning provides a general framework for modeling complex relations in transportation systems. As such, deep learning frameworks are well-suited to many optimization problems in transportation. This paper presents an innovative deep learning architecture for applying reinforcement learning and calibrating a transportation model. We have demonstrated, deep learning is a viable option compared to other metamodel based approaches. Our calibration and reinforcement learning examples demonstrate how to develop and apply deep learning models in transportation modeling. 

At the same time, there are significant challenges associated with using deep learning for optimization problems. Most notably, the issue of performance of deep reinforcement learning~\cite{wu2018variance}. Though theoretical bounds on performance of different RL algorithms do exist, the research done over the past few decades showed that worst case analysis is not the right framework for studying artificial intelligence: every model that is interesting enough to use in practice leads to computationally hard problems \cite{moitra13}. Similarly, while there are many important theoretical results  that show very slow convergence of many RL algorithm, it was shown to work well empirically on specific classes of problems. The convergence analysis developed for RL techniques is usually asymptotic and worst case.  Asymptotic optimality was shown by \cite{watkins_q-learning_1992} who shows  that $Q$-learning, which is an iterative scheme to learn optimal policies, does converge to optimal solution $Q^*$ asymptotically. Littman et.el. \cite{littman_generalized_1996} showed that a general reinforcement learning models based on exploration model does converge to an optimal solution. It is not uncommon for convergence rates in practice to be much better than predicted by worst case scenario analysis. Some of the recent work suggests that using recurrent architectures for Value Iteration Networks (VIN) can achieve good empirical performance compared to fully connected architectures~\cite{lee2018lstm}. Adaptive approaches that rely on meta-learning were shown to improve performance of reinforcement learning algorithms~\cite{al2017continuous}.

Another issue that requires further research is the bias-variance trade-off in he context of deep reinforcement learning. Traditional regularization techniques that add stochasticity to RL functions do not prevent from over-fitting~\cite{2018arXiv180406893Z}.

In the meantime, deep learning and deep reinforcement learning are likely to exert greater and greater influence in the practice of transportation.

\ifarxiv
\bibliographystyle{plain}
\else
\bibliographystyle{IEEEtran}
\fi
\bibliography{paper_bib}

\end{document}